\documentclass{vldb}
\usepackage{graphicx}

\begin{document}


\title{A Homogeneous Reaction Rule Language for \\
Complex Event Processing}


%
%
%
%

\numberofauthors{3}

\author{
%
%
\alignauthor Adrian Paschke\\
       \affaddr{RuleML Inc., Canada}\\
       \email{adrian.paschke AT gmx.de}
\alignauthor Alexander Kozlenkov\\
       \affaddr{Betfair Ltd., London}\\
       \email{alex.kozlenkov AT betfair.com}
\alignauthor Harold Boley\\
       \affaddr{National Research Council, Canada}\\
       \email{harold.boley AT nrc.gc.ca}
}

\maketitle

\begin{small}

\begin{abstract}
Event-driven automation of reactive functionalities for complex event processing is an urgent need in today's distributed
service-oriented architectures and Web-based event-driven environments. An important problem to be addressed is how to correctly and efficiently
capture and process the event-based behavioral, reactive logic embodied in reaction rules, and combining this with other conditional decision
logic embodied, e.g., in derivation rules. This paper elaborates a homogeneous integration approach that combines derivation
rules, reaction rules and other rule types such as integrity constraints into the general framework of logic programming, the
industrial-strength version of declarative programming. We describe syntax and semantics of the language, implement a distributed
web-based middleware using enterprise service technologies and illustrate its adequacy in terms of expressiveness, efficiency and scalability through
examples extracted from industrial use cases. The developed reaction rule language provides expressive features such as modular ID-based updates
with support for external imports and self-updates of the intensional and extensional knowledge bases, transactions including integrity testing
and roll-backs of update transition paths. It also supports distributed complex event processing, event messaging and event querying via
efficient and scalable enterprise middleware technologies and event/action reasoning based on an event/action algebra implemented by an
interval-based event calculus variant as a logic inference formalism.
\end{abstract}

\section{Introduction}
Event-driven applications based on reactive rules which trigger actions as a response to the detection of events have been extensively studied
during the 1990s. Stemming from the early days of programming language where system events were used for interrupt and exception handling,
active event-driven rules have received great attention in different areas such as active databases which started in the late 1980s, real-time
applications and system and network management tools which emerged in the early 1990s as well as publish-subscribe systems which appeared in the
late 1990s. Recently, there has been an increased interest in industry and academia in event-driven mechanisms and high-level Event-Driven
Architectures (EDA). (Pro-)active real-time or just-in-time reactions to events are a key factor in upcoming agile and flexible IT
infrastructures, distributed loosely coupled service oriented environments or new business models such as On-Demand or Utility computing.
Industry trends such as Real-Time Enterprise (RTE), Business Activity Management (BAM) or Business Performance Management (BPM) and closely
related areas such as Service Level Management (SLM) with monitoring and enforcing Service Level Agreements (SLAs)
\cite{Paschke2006,Paschke2007} are business drivers for this renewed interest. Another strong demand for event processing functionalities comes
from the web community, in particular in the area of Semantic Web and Rule Markup Languages (e.g. RuleML \cite{Boley2006}).

Different rule-based approaches to reactive event processing have been developed, which have for the most part proceeded separately and have led
to different views and terminologies:

\begin{itemize}
    \item Active databases in their attempt to combine techniques from expert systems and databases to support automatic triggering of global rules in
    response to events and to monitor state changes in database systems have intensively explored and developed the ECA paradigm and event algebras
    to compute complex events and trigger reactions according to global ECA rules.
    \item Event notification and messaging systems facilitate the communication of events in a distributed environment (push/pull) or monitor external systems which notify subscribed clients upon detected events. Typically, the interest here
    is in a context-dependent event sequence which follows e.g. a communication protocol or coordination workflow, rather than in single
    event occurrences which trigger immediate reactions.
    \item In event/action logics, which have their origins in the area of knowledge representation (KR) and logic programming (LP), the focus is on the formalization of action/event axioms and on the inferences that can be made from the happened or planned
    events/actions.
\end{itemize}

For a classification of the event/action/state processing space and a survey on different reaction rules approaches see \cite{Paschke2006a}.

In this paper we present an approach based on logic programming which combines and exploit the advantages of the different logic and rule-based
approaches and implement a homogeneous reaction rule language and an efficient scalable middleware which supports, e.g., complex event- and
action processing, event communication/messaging, formalization of reaction rules in combination with other rule types such as derivation rules
and transactional ID-based knowledge updates which dynamically change the intensional and extensional knowledge base (KB). We further show how
this approach fits into RuleML, a standardization initiative for describing different rule types on the (Semantic) Web.

In section 2 we introduce global reactions rules which follow the ECA paradigm. In section 3 we develop an interval-based event algebra which
supports logic-based complex event processing in three phases: event definition, event selection (detection/querying) and event consumption (via
rule-based consumption policies). In section 4 we introduce transactional module-based update actions into our reaction rule language which are
the basis for highly-distributed open and modular web-based ordered logic programs (OLPs). Section 5 describes Reaction RuleML which serves as
a platform independent rule interchange format and XML serialization language. In section 6 we implement a highly scalable and efficient
communication and management middleware which allows deploying rule inference services for rule-based complex event processing and
service-oriented computing on the Web. Section 7 concludes this paper with a discussion of our rule-based design artifact for application
domains such as IT Service Management (ITSM), Business Activity Monitoring (BAM) and Business Process Management (BPM), and Service Oriented
Computing.

\section{Global Reaction Rules}
Global reaction rules typically follow the Event-Condition-Action (ECA) paradigm: "\textit{on Event and Condition do Action}". We have
implemented a tight integration of global ECA rules and derivation rules. Both rule types are represented together in a homogeneous knowledge
base (KB): $KB = <R, E, F, I>$, where $R$ is the set of derivation rules, $E$ the set of ECA rules, $F$ the set of facts and $I$ the set of
integrity constraints. The KB might be updated over time, i.e. the intensional and extensional knowledge might be changed dynamically during
runtime incorporating self-updates (add/remove) triggered by active rules or imports of external modules. The approach is based on the logic
programming paradigm where the KB is an extended logic program (ELP) with finite function nestings, variables, (non-monotonic) default negation and
explicit negation, a typed logic supporting Semantic Web types from external ontologies (RDFS/OWL) \cite{Paschke2007,Paschke2006b} and external
Java types and procedural attachments using Java objects/methods in rule derivations \cite{Kozlenkov2004,Kozlenkov2007}. In the following we use
the standard LP notation with an ISO Prolog related scripting syntax called Prova \cite{Kozlenkov2004,Kozlenkov2007}. We assume that the reader
is familiar with logic programming techniques \cite{Lloyd1987}.

\subsection{Syntax of global ECA rules}
A global reaction rule is formalized as an extended ECA rule, represented in the KB as a 6-ary fact with the reserved predicate name $eca$:
$eca(T,E,C,A,P,EL)$, where $T$ (time), $E$ (event), $C$ (condition),  $A$ (action), $P$ (post condition), $EL(se)$ are complex terms/functions
[10]. The complex terms are meta-interpreted by the ECA interpreter as (sub)goals on derivation rules in the KB which are used to implement
the functionality of each part of an ECA rule. That is, the full expressiveness of derivation rules in extended LPs with logical connectives,
variables, finite function nestings, (non-monotonic) default and explicit negation as well as linear sequential operators such as "cuts", serial updates
and procedural attachments can be used to declaratively implement arbitrary complex behavioural reaction logic, whereas the core ECA rules'
syntax stays compact. The implemented global derivation rules can be reused several times in different ECA rules, leading to a compact
homogenous rule base (modularization).

The time part ($T$) of an ECA rule defines a pre-condition (an explicitly stated temporal event defining a validity time or event processing
window) which specifies a specific point or window of time at which the ECA rule should be processed by the ECA processor, either absolutely
(e.g., "at 1 o'clock on the 1st of May 2006), relatively (e.g., 1 minute after event X was detected) or periodically ("e.g., "every 10
seconds"). The post-condition ($P$) is evaluated after the action has been executed. It might be used to prevent backtracking from different
variable bindings via cuts or it might be used to apply post-conditional integrity and verification/validation tests in order to safeguard
transactional knowledge updates in ECA rules. The else part ($EL$) defines an alternative action which is executed in case the ECA rule could
not be completely executed. This is in particular useful to specify default (re-)actions or trigger exceptional failure handling policies. The
parts of a reaction rule might be blank, i.e. always satisfied, stated with "\_", e.g., $eca(t, e, \_, a,\_,\_)$. Blank parts might be
completely omitted leading to specific types of reactive rules, e.g. standard ECA rules (ECA: $eca(e,c,a)$ ), production style rules (CA:
$eca(c,a)$ ), extended ECA rules with post conditions (ECAP: $eca(e,c,a, p)$ ). Variable bindings to ground knowledge which is derived from the
queried derivation rules or the queried external data sources (e.g. a relational database or constructive procedural function/method) are
supported together with unification and backtracking as in logic programming, e.g., $eca(detect(e1(Context),Time),\_,fire(a1(Context),Time))$.
This is in particular useful to interchange context information, e.g. between the event and the action or the condition and the action part.
Negation is supported, e.g., $eca(not(ping(service1)),neg($\\
$maintenance(service1)),create(troubleticket(service1)))$. The\\ event of this ECA rule is detected if the ping (might be an external web
service call) fails by default, i.e. the (invoke) call on the external procedural function fails or no answer is received (i.e. the information
is missing). If the condition can be explicitly proved as false, i.e. $service1$ is definitely not in maintenance state, a trouble ticket for
this unavailable service is automatically created.

\begin{scriptsize}
\textbf{Example}: \textit{Every 10 seconds it is checked (time) whether there is a service request by a customer (event). If there is a service
request a list of all currently unloaded servers is created (condition) and the service is loaded to the first server (ac-tion). In case this
action fails, the system will backtrack and try to load the service to the next server in the list. Otherwise it succeeds and further
backtracking is prevented (post-condition cut) . If no unloaded server can be found, the else action is triggered, sending a notification to the
customer.}
\begin{verbatim}
eca(
    every10Sec(), % time
    detect(request(Customer, Service),T), % event
    find(Server), % condition
    load(Server, Service), % action
    !, % postcondition
    notify(Customer, "Service request temporarily rejected").
).
% time
every10Sec() :-  sysTime(T), interval( timespan(0,0,0,10),T).
% event
detect(request(Customer, Service),T):-
    occurs(request(Customer,Service),T),
    consume(request(Customer,Service)).
% condition
find(Server) :- sysTime(T), holdsAt(status(Server, unloaded),T).
% action
load(Server, Service) :-
    sysTime(T),
    rbsla.utils.WebService.load(Server,Service),%proc. attachment
    add(key(Server), "happens(loading(_0),_1).",  [Server, T]).
% alternative action
notify(Customer, Message):-
        sendMessage(Customer, Message).
\end{verbatim}
\end{scriptsize}

The state of each server might be managed via an Event Calculus (EC) formalization:

\begin{scriptsize}
\begin{verbatim}
terminates(loading(Server),status(Server,unloaded),T).
initiates(unloading(Server),status(Server,unloaded),T).
\end{verbatim}
\end{scriptsize}

The example includes possible backtracking to different variable bindings. In the condition part all servers that are in the state $unloaded$
are bound to the variable $Server$. If the action which tries to load a server with the service succeeds, further backtracking is prevented by
the post-condition cut. If no unloaded server can be found for the customer request, the "else" action is executed which notifies the
customer.

\subsection{Declarative Semantics of Global ECA Rules}

The declarative semantics of ECA rules is directly inherited from the semantics of the underlying rule/inference system. The goals/queries
defined by the (truth-valued model-theoretic) functions denoting the ECA rule parts are actively used to query the KB and evaluate the
derivation rules (or external Boolean-valued function implementations) that implement the functionality of the ECA rule parts.

\begin{scriptsize}
\textbf{Definition}: \textit{An extended ECA rule is interpreted as a conjunction of (sub)goals (its complex terms) which must be processed in a
left-to-right order starting with the goal denoting the time part, in order to capture the forward-directed semantics of an ECA rule: $ECA_{i} =
T \wedge E \wedge ((C \wedge A \wedge P) \vee EL)$, where $ECA$ is the top goal/query which consists of the subgoals $T$,$E$,$C$,$A$,$P$,$EL$.
An ECA rule succeeds, i.e. $ECA_{i}$ is entailed by the KB, if the subgoals succeed: $KB \models ECA_{i}$ iff $KB \models (\forall \overline{X})
(T \wedge E \wedge ((C \wedge A \wedge P) \vee EL))$, where $\overline{X}$ is the set of variables occurring free in $ECA_{i}$.}
\end{scriptsize}

The post-condition acts as a constraint on the KB state after the action has been performed. In particular, actions with effects on the KB such
as knowledge updates which transit the current KB state to the next state can be tested by integrity constraints. If the integrity tests fail,
transactional knowledge updates are rolled back according to the semantics of the transaction logic in our approach. In case of external
actions, compensating actions can be called in the else part, if the external system provides appropriate API methods which support
transactions. That is, the action part $A$ of a reaction rule can lead to overall rule success only if, besides $success(A)$, the (pre)condition
before the action and the postcondition after the action are evaluated to true: $\forall \overline{X} (C \wedge success(A) \wedge P)$. For a
detailed description of the declarative semantics, in particular the implemented extended well-founded semantics for ELPs see
\cite{Paschke2007}.

\subsection{Operational Semantics of Global ECA Rules}

In order to integrate the (re)active behavior of ECA rules into goal-driven backward-reasoning, the goals defined by the complex terms in the ECA
rules are meta-interpreted by an additional ECA interpreter. The interpreter implements the forward-directed operational semantics of the ECA
paradigm. The ECA interpreter provides a general wrapper interface which can be specialized to a particular query API of an arbitrary
backward-reasoning inference engine. This means, the ECA meta-interpreter is used as a general add-on attached to an LP system (a derivation rule
engine) extending it with reasoning and processing features for reactive rules. The task of processing an ECA rule by querying the respective
derivation rules using the defined complex terms in an ECA rule as queries on the KB is solved by a daemon (implemented within the ECA
interpreter). The daemon is a kind of adapter that frequently issues queries on the ECA rules in order to simulate the active behavior in
passive goal-driven LP systems. Proof-theoretically, it applies the ECA subgoals of a top query formed by an ECA rule one by one on the
KB (the current KB state) using the inference rules of the underlying backward-reasoning inference engine to deductively prove the syntactic
derivability from the clauses in the KB. The process is as follows:

\begin{scriptsize}
\begin{enumerate}
    \item it queries (repeatedly -- in order to capture
    updates to reactive rules) the KB and derives all ECA rules
    represented in the KB by the universal query $eca(T,E,C,A,P,EL)?$,
    \item it adds the derived ECA rules to its internal active KB,
    which is a kind of volatile storage for reactive rules and
    temporal event data, and
    \item finally, it processes the ECA rules sequentially or in parallel, depending on the configuration, using a thread pool.
\end{enumerate}
\end{scriptsize}

The forward-directed execution of the ECA paradigm is given by the strictly positional order of the terms in the ECA rules. That is, first the time
part is queried/evaluated by the ECA processor (daemon), when it succeeds then the event part is evaluated, then the condition, and so on. The
computed (ground) substitutions $\theta$ of the variables for each subgoal in a rule $ECA_{i}$ are unified by the ECA interpreter with their
variable variants in the subsequent subgoals of the top ECA query. The interpreter also implements the common LP backtracking mechanism to
backtrack to different variable bindings. In order to enable parallel processing of ECA rules, the ECA processor implements a thread pool where
each ECA rule is executed in a separate thread, if its time part succeeds.

\begin{scriptsize}
\textbf{Example}: "\textit{Every 10 seconds it is checked (time) whether there is an incoming request by a customer to book a flight to a
certain destination (event). Whenever this event is detected, a database look-up selects a list of all flights to this destination (condition)
and tries to book the first flight (action). In case this action fails, the system will backtrack and book the next flight in the list; otherwise
it succeeds (post-condition cut) sending a "flight booked" notification. If no flight can be found to this destination, i.e. the condition fails,
the else action is triggered, sending a "booked up" notification back to the customer.}"
\begin{verbatim}
eca( every10Sec(), detect(request(Customer, Destination),T),
    find(Destination, Flight), book(Customer, Flight), !,
    notify(Customer, bookedUp(Destination) ).
% time derivation rule
every10Sec() :-  sysTime(T), interval( timespan(0,0,0,10),T).
% event derivation rule
detect(request(Customer, FlightDestination),T):-
    occurs(request(Customer,FlightDestination),T),
    consume(request(Customer,FlightDestination)).
% condition derivation rule
find(Destination,Flight) :-
    on_exception(java.sql.SQLException,on_db_exception()),
    dbopen("flights",DB), sql_select(DB,"flights", [flight,
    Flight], [where, dest=Destination"]).
% action derivation rule
book(Cust, Flight) :-
    flight.BookingSystem.book(Flight, Cust),
    notify(Cust,flightBooked(Flight)).
% alternative action derivation rule
notify(Customer, Message):- sendMessage(Customer, Message).
\end{verbatim}
\end{scriptsize}

If the action succeeds further backtracking is prevented by the post-condition cut "!". If no flight can be found for the customer request,
the else action is executed which notifies the customer about this. The condition derivation rule accesses an external data source via an SQL
query to a relational data base.\\
\\
The homogeneous combination of global ECA-style reaction rules and derivation rules within the same representation language paves the way to
the (re-)use of various other useful logic formalisms in reactive rules, such as procedural attachments, defeasible rules with rule priorities for
conflict resolution, transactional update actions, event/action logics for temporal event calculations and complex event/action reasoning. Moreover, it
relates ECA rules to other rule types such as integrity constraints or normative rules. As a result, the high expressive power and the clear
logical semantics of these formalisms is also accessible to reaction rules.

\section{Interval-based Event Calculus Event / Action Algebra}

Events and actions in reactive ECA rules are typically not atomic but complex, consisting of several atomic events or actions that must
occur or be performed in the defined order and quantity so as to detect complex events or execute complex actions, e.g. an ordered sequence
of events or actions. This topic has been extensively studied in the context of active databases and event algebras, which provide the operators
to define complex event types.

As we have pointed out in \cite{Paschke2005c} typical event algebras of active database systems for complex event definitions such as Snoop
\cite{Chakravarthy1994} show inconsistencies and irregularities in their operators. For instance, consider the sequence $B;(A;C)$ in Snoop. The
complex event is detected if $A$ occurs first, and then $B$ followed by $C$, i.e. an event instance sequence (EIS) $EIS={a, b, c}$ will lead to
the detection of the complex event, because the complex event $(A;C)$ is detected with associated detection time of the terminating event $c$
and accordingly the event $b$ occurs before the detected complex event $(A;C)$. However, this is not the intended semantics: Only the sequence
$EIS={b, a, c}$ should lead to the detection of the complex event defined by $B;(A;C)$. In the semantics of Snoop, which uses the detection time
of the terminating event as occurrence time of a complex event, both complex event definitions $B;(A;C)$ and $A;(B;C)$ are equal. This problem
arises from the fact that the events, in the active database sense, are simply detected and treated as if they occur at an atomic instant, in
contrast to the durative complex events, in the KR event/action logics sense, which occur over an extended interval.

To overcome such unintended semantics and provide verifiable and traceable complex event computations (resp. complex actions) we have
implemented an interval-based Event Calculus (EC) variant and refined the typical event algebra operators based on it. In the interval-based
Event Calculus \cite{Paschke2005c} all events are regarded to occur in a time interval, i.e. an event interval $[e1,e2]$ occurs during the time
interval $[t1,t2]$ where $t1$ is the occurrence time of $e1$ and $t2$ is the occurrence time of $e2$. In particular, an atomic event occurs in
the interval $[t,t]$, where $t$ is the occurrence time of the atomic event.

The interval-based EC axioms describe when events / actions \emph{occur} (transient consumption view), \emph{happen} (non-transient reasoning
view) or are \emph{planned} to happen (future abductive view) within the EC time structure, and which properties (\emph{fluents} or \textit{event
intervals}) are \emph{initiated} and/or \emph{terminated} by these events under various circumstances.

\begin{scriptsize}
\textbf{Definition}: \textit{(\textbf{Interval-based Event Calculus Language}) An interval-based EC signature is a multi-sorted signature with
equality, with a sort $E$ for events, a sort $F$ for fluents (states or event intervals $[E1,E2]$), and a sort $T$ for timepoints. An EC
language $\Sigma^{EC}$ is a tuple $\langle \overline{E},\overline{F},\overline{T},\leq \rangle$ where $\leq$ is a partial ordering defined over
the non-empty set $\overline{T}$ of time points, $\overline{E}$ is a non-empty set of events/actions and $\overline{F}$ is a non-empty set of
fluents. Timepoints, events/actions and fluents are n-ary functional literals $L$ or $\neg L$ which might be reified typed functions or external
object instantiations, e.g. constructive views over relational databases storing the event occurrences, Java object instances, or XML data such
as Common Base Event, or Semantic Web event class instances in RDF or OWL.} \end{scriptsize}

The basic $holdsAt$ axiom of the classical EC \cite{Kowalski1986} for temporal reasoning about fluents is redefined in the interval-based EC to
the axiom $holdsInterval([E1,E2],[T1,T2])$ to capture the semantics of event intervals which hold between a time interval:

\begin{scriptsize}
\begin{verbatim}
holdsInterval([E1,E2],[T11,T22]):-
    event([E1],[T11,T12]), event([E2],[T21,T22]),
    [T11,T12]<=[T21,T22], not(broken(T12,[E1,E2],T21).
\end{verbatim}
\end{scriptsize}

The event function $event([Event],[Interval])$ is a meta-function to translate instantaneous event occurrences into interval-based events:
$event([E],[T,T]) :- occurs(E,T)$. It is also used in the event algebra meta-program to compute complex events from occurred raw events
according to their event type definitions. The $broken$ function tests whether the event interval is not broken between the the initiator event
and the terminator event by any other explicitly specified terminating event:

\begin{scriptsize}
\begin{verbatim}
broken(T1,Interval,T2):-
          terminates(Terminator,Interval,[T1,T2]),
          event([Terminator],[T11,T12]), T1<T11, T12<T2.
\end{verbatim}
\end{scriptsize}

The declarative semantics is given by interpretations which map event intervals $[E1,E2]$ and time intervals $[T1,T2]$ to truth values.

\begin{scriptsize}
\textbf{Definition}: \textit{(\textbf{Interval-based Event Calculus Interpretation}) An interpretation is a mapping $I: [T1,T2]\,x\,[E1,E2] \mapsto
\{true, false\}$.}\\
\textbf{Definition}: \textit{(\textbf{Event Calculus Satisfaction}) An interpretation $I$ satisfies an event interval $[E1,E2]$ at a time interval $[T1,T2]$ if $I([E1,E2],$\\ $[T1,T2]) = true$ and $I(\neg [E1,E2],[T1,T2])=false$.}\\
\textbf{Definition}: \textit{(\textbf{Instantiation and Termination}) Let $\Sigma^{EC}$ be an interval-based EC language, $D^{EC}$ be a domain
description (an EC program) in $\Sigma^{EC}$ and $I$ be an interpretation of $\Sigma^{EC}$. Then an event interval $[E1,E2]$ is instantiated at
time point $T1$ in $I$ iff there is an event $E1$ such that there is a statement in $D^{EC}$ of the form $occurs(E1,T1)$ and a statement in
$D^{EC}$ of the form $initiates(E1,[E1,E2],T)$. A event interval $[E1,E2]$ is terminated at time point $T2$ in $I$ iff there is an event $E2$
such that there is a statement in $D^{EC}$ of the form $occurs(E2,T2)$ and a statement in $D^{EC}$ of the form $terminates(E2,$\\ $[E1,E2],T)$.}
\end{scriptsize}

An interpretation qualifies as a model for a given domain description, if:

\begin{scriptsize}
\textbf{Definition}: \textit{(\textbf{Event Calculus Model}) Let $\Sigma^{EC}$ be an interval-based EC language, $D^{EC}$ be a domain
description in $\Sigma^{EC}$. An interpretation $I$ of $\Sigma^{EC}$ is a model of $D^{EC}$ iff $\forall [E1,E2] \in \overline{F}$ and $T1 \leq
T2 \leq T3$ the following holds:}

\begin{enumerate}
    \item If $[E1,E2]$ has not been instantiated or terminated at
    $T2$ in $I$ wrt $D^{EC}$ then $I([E1,E2],[T1,T1]) = I([E1,E2],[T3,T3])$
    \item If $[E1,E2]$ is initiated at $T1$ in $I$ wrt $D^{EC}$, and not
    terminated at $T2$ the $I([E1,E2],[T3,T3])=true$
    \item If $[E1,E2]$ is terminated at $T1$ in $I$ wrt $D^{EC}$ and not
    initiated at $T2$ then $I([E1,E2],[T3,T3])=false$
\end{enumerate}
\end{scriptsize}

The three conditions define the persistence of complex event intervals as time progresses. That is, only events/actions have an effect on the
changeable event interval states (condition 1) and the truth value of a complex event state persists until it has been explicitly changed by
another terminating event/action (condition 2 and 3). A domain description is consistent if it has a model. We now define entailment wrt to the
meta-program domain description which we have implemented as a logic program:

\begin{scriptsize}
\textbf{Definition}: \textit{(\textbf{Event Calculus Entailment}) Let $D^{EC}$ be an interval-based EC domain description. A event interval
$[E1,E2]$ holds at a time interval $[1,T2]T$ wrt to $D^{EC}$, written $D^{EC} \models holdsInterval($\\ $[E1,E2],[T1,T2])$, iff for every
interpretation $I$ of $D^{EC}$, $I([E1,E2],$\\ $[T1,T2])=true$. $D^{EC} \models neg(holdsInterval([E1,E2],[T1,T2]))$ iff
$I([E1,E2],[T1,T2])=false$.}
\end{scriptsize}\\
\\
\textbf{Example}
\begin{scriptsize}
\begin{verbatim}
occurs(a,datetime(2005,1,1,0,0,1)).
occurs(b,datetime(2005,1,1,0,0,10)).
Query: holdsInterval([a,b],Interval)?
Result: Interval=
        [datetime(2005,1,1,0,0,1), datetime(2005,1,1,0,0,10)]
\end{verbatim}
\end{scriptsize}

Based on this interval-based event logics formalism, we now redefine the typical (SNOOP) event algebra operators and treat complex events resp.
actions as occurring over an interval rather than in terms of their instantaneous detection times. In short, the basic idea is to split the
occurrence interval of a complex event into smaller intervals in which all required component events occur, which leads to the definition of
event type patterns in terms of interval-based event detection conditions, e.g. the sequence operator ($;$) is formalized as follows $(A;B;C)$:

\begin{scriptsize}
\begin{verbatim}
detect(e,[T1,T3]):-
    holdsInterval([a,b],[T1,T2],[a,b,c]),
    holdsInterval([b,c],[T2,T3],[a,b,c]),
    [T1,T2]<=[T2,T3].
\end{verbatim}
\end{scriptsize}

For the complete formalization of the interval-based EC event algebra see the ContractLog KR \cite{Paschke2007,Paschke2006}. In order to make
definitions of complex events in terms of event algebra operators more comfortable and remove the burden of defining all interval conditions for
a particular complex event type as described above, we have implemented a meta-program which implements an interval-based EC event algebra in
terms of typical event operators:

\begin{scriptsize}
\begin{verbatim}
Sequence:           sequence(E1,E2, .., En)
Disjunction:        or(E1,E2, .. , En)
Mutual exclusive:   xor(E1,E2,..,En)
Conjunction:        and(E1,E2,..,En)
Simultaneous:       concurrent(E1,E2,..,En)
Negation:           neg([ET1,..,ETn],[E1,E2])
Quantification:     any(n,E)
Aperiodic:          aperiodic(E,[E1,E2])
\end{verbatim}
\end{scriptsize}

In order to reuse detected complex events in rules, e.g. in ECA rules or other complex events, they need to be remembered until they are
consumed, i.e. the contributing component events of a detected complex event should be consumed after detection of a complex event. This can be
achieved via the ID-based update primitives which allow adding or removing knowledge from the KB. We use these update primitives to add detected
event occurrences as new transient facts to the KB and consume events which have contributed to the detection of the complex event.

\textbf{Example}
\begin{scriptsize}
\begin{verbatim}
detect(e,T):-
    event(sequence(a,b),T), % detection condition
    add(eis(e), "occurs(e,_0).", [T]), % add e with key eis(e)
    consume(eis(a)), consume(eis(b)). % consume all a and b events
\end{verbatim}
\end{scriptsize}

If the detection conditions for the complex event $e$ are fulfilled, the occurrence of the detected event $e$ is added to the KB with the key
$eis(e)$ (eis = event instance sequence). Then all events that belong to the type specific event instance sequences of type $a$ and type $b$ are
consumed using their ids $eis(a)$ resp. $eis(b)$. Different consumption policies are supported such as "\textit{remove all events which belong
to a particular type specific eis}" or "\textit{remove the first resp. the last event in the eis}". If no consume predicate is specified in a
detection rule, the events are reused in the detection of other complex events. The detection rule of a complex event might be used in an ECA
rule to trigger the ECA rule.

For space reasons we have only discussed the processing of complex events, but the interval-based event calculus can also used for the
definition of complex actions, e.g. to define an ordered sequence of action executions or concurrent actions (actions which must be performed in
parallel within a time interval), with a declarative semantics for possibly required rollbacks, as defined above. For more information see
\cite{Paschke2005c,Paschke2007}.

\section{Event Messaging Reaction Rules}
The ECA rules described in the previous section are defined globally. Such global ECA rules are best suited to represent reaction rules that
actively detect or query internal and external events in a global context and trigger reactions. For instance, to actively monitor an external
system, data source, or service and to trigger a reaction whenever the system/service becomes unavailable. In a distributed environment with
independent system nodes that communicate with each other relative to a certain context (e.g. a workflow, conversation protocol state or complex
event situation), event processing requires event notification and communication mechanisms, and often needs to be done in a local context, e.g.
a conversation state or (business) process workflow. Systems either communicate events according to a predefined or negotiated
communication/coordination protocol or they subscribe to specific event types on a server (publish-subscribe). In the latter case, the server
monitors its environment and upon detecting an atomic or complex event (situation), notifies the concerned clients. Complex events may
correspond to pre-defined protocols or be based on event algebras including time-restricted sequences and conjunctions/disjunctions, which
permits events like \textit{A} occurs more than \textit{t} time units after \textit{B} to be expressed.

\subsection{Syntax of Event Messaging Reaction Rules}
In addition to global ECA-style reaction rules, we provide a homogenously-integrated event messaging reaction rule language, called Prova AA
\cite{Kozlenkov2006}, that includes constructs for asynchronously sending and receiving event messages via various communication protocols over
an enterprise service bus and for specifying reaction rules for processing inbound messages. These messaging reaction rules do not require
separate threads for handling multiple conversation situations simultaneously. The main language constructs are: \textit{sendMsg} predicates,
reaction \textit{rcvMsg} rules, and \textit{rcvMsg} or \textit{rcvMult} inline reactions:

\begin{scriptsize}
\begin{verbatim}
sendMsg(XID,Protocol,Agent,Performative,Payload |Context)
rcvMsg(XID,Protocol,From,queryref,Paylod|Context)
rcvMult(XID,Protocol,From,queryref,Paylod|Context)
\end{verbatim}
\end{scriptsize}

where \textit{XID} is the conversation identifier (conversation-id) of the conversation to which the message will belong. \textit{Protocol}
defines the communication protocol. More than 30 protocols such as JMS, HTTP, SOAP, Jade are supported by the underlying enterprise service bus
as efficient and scalable object-broker and communication middleware \cite{Mule2006}. \textit{Agent} denotes the target (an agent or service
wrapping an instance of a rule engine) of the message. \textit{Performative} describes the pragmatic context in which the message is send. A
standard nomenclature of performatives is e.g. the FIPA Agents Communication Language ACL or the BPEL activity vocabulary. \textit{Payload}
represents the message content sent in the message envelope. It can be a specific query or answer or a complex interchanged rule base (set of
rules and facts).

\textbf{Example:}
\begin{scriptsize}
\begin{verbatim}
% Upload a rule base read from File to the host
% at address Remote via JMS
upload_mobile_code(Remote,File) :-
    % Opening a file returns an instance
    % of java.io.BufferedReader in Reader
    fopen(File,Reader),
    Writer = java.io.StringWriter(),
    copy(Reader,Writer),
    Text = Writer.toString(),
    % SB will encapsulate the whole content of File
    SB = StringBuffer(Text),
    sendMsg(XID,jms,Remote,eval,consult(SB)).
\end{verbatim}
\end{scriptsize}

The example shows a reaction rule that sends a rule base from an external \textit{File} to the agent service \textit{Remote} using JMS as
transport protocol. The inline sendMsg reaction rules is locally used within a derivation rule, i.e. only applies in the context of the
derivation rule. The corresponding global receiving reaction rule could be:

\begin{scriptsize}
\begin{verbatim}
rcvMsg(XID,jms,Sender,eval,[Predicate|Args]):-
    derive([Predicate|Args]).
\end{verbatim}
\end{scriptsize}

This rule receives all incoming JMS based messages with the pragmatic context \textit{eval} and derives the message content. The list notation
$[Predicate \mid Args]$ will match with arbitrary n-ary predicate functions, i.e., it denotes a kind of restricted second order notation since
the variable \textit{Predicate} is always bound, but matches to all predicates in the signature of the language with an arbitrary number of
arguments \textit{Args}).

\subsection{Semantics of Event Messaging Reaction Rules}

The semantics treats the send and receive constructs as special built-in literals. The execution flow is stored in a transparently constructed
temporal reaction rule that is activated once the message matching the specified pattern in $rcvMsg$ has arrived. The engine then continues
processing other incoming messages and goals/queries. The temporal rule includes the pattern of the message specified in $rcvMsg$ (or $rcvMult$)
in the head of the rule and all remaining goals at the current execution point in its body. This behaviour is different from suspending the
current thread and waiting for the replies to arrive. The rule engine maintains one main thread while allowing an unlimited number of
conversations to proceed at the same time without incurring the penalty and limitations of multiple threads otherwise used solely for the
purpose of maintaining several conversation states at the same time.

The conversation and event context based semantics of global and local inline receive and send reaction rules allows implementing typical
semantics of state machines or workflow-style systems such as Petri nets or pi-calculus, which can be used for complex event processing and
workflow process executions (e.g. in style of BPEL). Figure \ref{PetriNet} shows an example of a event-based workflow which is implemented as
follows:
\begin{figure}
\centering
\includegraphics[width=8cm]{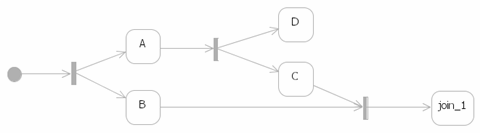}
\caption{Example Rule-Based Workflow} \label{PetriNet}
\end{figure}

\begin{scriptsize}
\begin{verbatim}
process_join() :-
    iam(Me),
    init_join(XID,join_1,[c(_),b(_)]),
    fork_a_b(Me,XID).
fork_a_b(Me,XID) :-
    rcvMsg(XID,self,Me,reply,a(1)),
    fork_c_d(Me,XID).
fork_a_b(Me,XID) :-
    rcvMsg(XID,self,Me,reply,b(1)),
    join(Me,XID,join_1,b(1)).
fork_c_d(Me,XID) :-
    rcvMsg(XID,self,Me,reply,c(1)),
    % Tell the join join_1 that a new pattern is ready
    join(Me,XID,join_1,c(1)).
% invoked by join once all the inputs are assembled.
join_1(Me,XID,Inputs) :-
    println(["Joined for XID=",XID," with inputs: ",Inputs]).
% Prints
% Joined for XID=agent@hostname001 with inputs [[b,1],[c,1]]
\end{verbatim}
\end{scriptsize}

The behaviour of the inbound links defined by the $rcvMsg$ reaction rules is similar to BPEL links. However, the declarative rule-based approach
provides a much more expressive declarative programming language to represent complex event processing logic, conditional reactions (activities)
and complex event-based workflow patterns such as Join, Simple Merge, Cancel Activity, Multi-Choice, Structured Loop, Deferred Choice and
Milestone (see \cite{Kozlenkov2007} for an implementation). The homogenous language approach allows arbitrary combinations of logic and
meta-logic programming formalisms with global ECA and context-specific messaging reaction rules.

The following example demonstrates the use of messaging reaction rules for detecting a complex event pattern that includes event sequencing,
disjunction, and event branching that depends on non-arrival of specified events: Event pattern is $A; (\neg B^{*} B$ $or$ \\ $\neg E^{*} C);
D$, where '$*$' denotes $0..*$ instances of the messages it follows.
\begin{scriptsize}
\begin{verbatim}
rcvMsg(XID,Protocol,Sender,inform,A) :-
      partition_id(ID1),  partition_id (ID2),
      detect_bc(XID,ID1,ID2),
      rcvMsg(XID,Protocol,Sender,inform,D),
      % inform the clients here about the detected situation
      ...
detect_bc(XID,ID1,ID2) :-
  rcvMsgP([ID1],[ID1],rcvMsg(XID,Protocol,Sender,inform,B)).
detect_bc(XID,ID1,ID2) :-
  rcvMsgP([ID1,ID2],[ID1],rcvMsg(XID,Protocol,Sender,inform,C)).
detect_bc(XID,ID1,ID2) :-
rcvMsgP([ID1],[ID2],rcvMsg(XID,Protocol,Sender,inform,E)), fail().
\end{verbatim}
\end{scriptsize}

In the example, message $A$ is the initiator of the complex event situation and has a reaction rule $rcvMsg$ beginning the situation detection.
This reaction rule is executed whenever a node receives a message with correlation-id $XID$ on $Protocol$ from $Sender$ with payload matching
the pattern $A$. The subsequent events contain two intersecting detection partitions ($ID1$ and $ID2$) that are active whenever the reactions to
corresponding events are enabled. For example, the reaction to event $C$ is only enabled while both partitions are active. The built-ins
$partition\_id$ generate unique id's for each partition. The rules for the predicate $detect\_bc$ spawn three reactions represented by the
built-in $rcvMsgP$ representing inline detection of events depending on inbound and outbound partition links. This behaviour is similar but
opposite to BPEL links. Outbound links (argument 2 of $rcvMsgP$) represent partitions that become inactive once the message matching the pattern
in argument 3 of $rcvMsgP$ is received. Inbound links (argument 1 of $rcvMsgP$) represent the partitions that must be active for the reaction to
be active. Inactive reactions are removed from the system. Consider now the $EIS \{A;E;B;D\}$. The initiator $A$ is detected by the first rule
and three reactions for events $B$, $C$ and $D$ are activated. The event $E$ deactivates the partition $ID2$ which removes the branches waiting
for the event $C$ and $E$. The event $B$ succeeds the first $detect\_bc$ rule while the last event $D$ is received by the built-in reaction
$rcvMsg$ that waits for this event skipping any events that do not match. The example demonstrates that the semantics of complex event detection
should fully respect the mutual ordering of events and also allow for flexibility in allowing or disallowing intervening events in conditional
branches. The messaging reaction rules can also easily detect conjunctions of events (not shown here due to the lack of space).

In summary, the messaging style reaction rules complement the global ECA rules. We will now illustrate the combination of active global ECA and
messaging reaction rules by a typical use case found in industry:

\begin{scriptsize}
\textbf{Example}: \textit{A Manager node is responsible for holding housekeeping information about various servers playing different roles. When
a server fails to send a heartbeat for a specified amount of time, the Manager assumes that the server failed and cooperates with the Agent
component running on an unloaded node to resurrect it. A reaction rule for receiving and updating the latest heartbeat in event notification
style is}:

\begin{verbatim}
rcvMsg(XID,Protocol,FromIP,inform,heartbeat(Role,RemoteTime)) :-
    time(LocalTime)
    update(key(FromIP,Role),"heartbeats(_0, _1, _2, _3).",
                   [ FromIP, Role, RemoteTime, LocalTime] ).
\end{verbatim}
\end{scriptsize}

The rule responds to a message pattern matching the one specified in the $rcvMsg$ arguments. \textit{XID} is the conversation-id of the incoming
message; \textit{inform} is the performative representing the pragmatic context of the message, in this case, a one-way information passed
between parties; \textit{heartbeat(...)} is the payload of the message. The body of the rule enquires about the current local time and updates
the record containing the latest heartbeat from the controller. This rule follows a push pattern where the event is pushed towards the rule
systems and the latter reacts. A pull-based global ECA rule that is activated every second by the rule engine and for each server that fails to
have sent heartbeats within the last second will detect server failures and respond to it by initiating failover to the first available unloaded
server. The accompanying derivation rules \textit{detect} and \textit{respond} are used for specific purpose of detecting the failure and
organizing the response.

\begin{scriptsize}
\begin{verbatim}
eca(
    every('1S') ,
    detect(controller_failure(IP,Role,'1S')) ,
    respond(controller_failure(IP,Role,'1S')) ) .

every('1S'):-
    sysTime(T),
    interval(timespan(0,0,0,1),T).

detect(controller_failure(IP,Role,Timeout)) :-
    sysTime(LocalTimeNow),
    heartbeats(IP,Role,RemoteTime,LocalTime),
    LocalTimeNow-LocalTime > Timeout.

respond(controller_failure(IP,Role,Timeout)) :-
    sysTime(LocalTime),
    first(holdsAt(status(Server,unloaded),LocalTime)),
    add(key(Server),
        "happens(loading(_0),_1).",[ Server, Local-Time]),
    sendMsg(XID,loopback,self,initiate,failover(Role,IP,Server)).
\end{verbatim}
\end{scriptsize}

The ECA logic involves possible backtracking so that all failed components will be resurrected. The state of each server is managed via an event
calculus formulation:

\begin{scriptsize}
\begin{verbatim}
initiates(loading(Server),status(Server,loaded),T).
terminates(unloading(Server),status(Server,loaded),T).
initiates(unloading(Server),status(Server,unloaded),T).
terminates(loading(Server),status(Server, loaded),T).
\end{verbatim}
\end{scriptsize}

The current state of each server is derived from the happened loading and unloading events and used in the ECA rule to detect the first server
which is in state \textit{unloaded}. This EC based formalization can be easily extended, e.g. with new states such as a maintenance state which
terminates an unloaded state, but is not allowed in case a server is already loaded:

\begin{scriptsize}
\begin{verbatim}
initiates(maintaining(Server),status(Server,maintenance),T):-
    not(holdsAt(status(Server,loaded),T)).
terminates(maintaining(Server),status(Server,unloaded),T).
\end{verbatim}
\end{scriptsize}

In summary, the messaging style reaction rules complement the global ECA rules in distributed coordination situations such as process workflows
and for distributed complex event processing.

\section{Transactional Module-based Updates}
Our reaction rule language provides support for three special update functions $add$, $remove$ and $transaction$ and two auxiliary functions
$commit$ and $rollback$. These update primitives are more expressive than the simple assert/retract primitives found in typical LP interpreters
such as Prolog and allow transactional as well as bulk updates of knowledge including updating of facts and rules. They enable arbitrary
knowledge updates, e.g., adding (\textit{add}) /removing (\textit{remove}) rules or complete rule sets including the import of knowledge from
external sources and transactional update (\textit{transaction}) which are rolled back if the execution fails. Transactions might be explicit
committed or rolledback. Each update has a unique ID with which it is asserted into the KB as a module, i.e. the KB is an ordered LP (OLP)
consisting of arbitrary nested and possibly distributed modules (rule sets) managed by their unique ID. Preferences with e.g. default rules and
priorities might be defined between rules and complete modules by an expressive integrity-preserving defeasible logic. Constructive scopes can
be used to constrain the level of generality of queries and explicitly close of open distributed KBs to support non-monotonic properties, e.g.
for negation-as-failure. For a detailed description of these expressive logic formalisms see the ContractLog KR
\cite{Paschke2007,Paschke2006}.

\begin{scriptsize}
\begin{verbatim}
add("./examples/test/test.prova")  % add an external script
add("http://rbsla.com/ContractLog/datetime.prova") % from URL
add(id1,"r(1):-f(1). f(1).")% add rule and fact
add(id2,"r(X):-f(X).")         % add rule "r(X):-f(X)."
p(X,Y) :- % object/variable place holders _N: _0=X ; _1=Y.
    add(id3,"r(_0):-f(_0), g(_0). f(_0). g(_1).",[X,Y]).
remove(id1)            % remove all updates with id
remove("./examples/test/test.prova")   % remove external update
\end{verbatim}
\end{scriptsize}

The examples show different variants of updates with external modules imported from their URIs, user-defined updates asserting rules and facts
and updates with previously bound variables from other goals which are integrated into the updates using place holders  $\_X$.

Remarkably, updates to the KB are handled as modules, i.e., as (smaller) logic programs which might contain further updates and imports of other
modules, leading to nested updates with hierarchical submodules. External LP scripts can be dynamically added and removed from the knowledge
base using their module object ids (typically the URI or an user defined label).

The semantics of updates in ContractLog is based on the notion of knowledge states and transitions from one state to another.

\begin{scriptsize}
\textbf{Definition}: \textit{(\textbf{Positive Update Transition}) A positive update transition, or simply positive update, to a knowledge state
$KB_{k}$ is defined as a finite set $U^{pos}_{oid}:=\{r_{N}:H \leftarrow B, fact_{M}: A \leftarrow \}$ with $A$ an atom denoting a fact, $H
\leftarrow B$ a rule, $N=0,..,n$ and $M=0,..m$ and $oid$ being the update oid which is also used as module oid to manage the knowledge as a new
module in the KB. Applying $U^{pos}_{oid}$ to $KB_{k}$ leads to the extended state $KB_{k+1} = \{KB_{k} \cup U^{pos}_{oid}\}$. Applying several
positive updates as an increasing finite sequence $U^{pos}_{oid_{j}}$ with $j=0,..,k$ and $U^{pos}_{oid_{0}}:=\emptyset$ to $KB_{0}$ leads to a
state $KB_{k} = \{KB_{0} \cup U^{pos}_{oid_{0}} \cup U^{pos}_{oid_{1}} \cup ... \cup U^{pos}_{oid_{k}}\}$.} \end{scriptsize}

Likewise, we define a \emph{negative update transition} as follows:

\begin{scriptsize}
\textbf{Definition}: \textit{(\textbf{Negative Update Transition}) A negative update transition, or for short a negative update, to a knowledge
state $KB_{k}$ is a finite set $U^{neg}_{oid}:=\{r_{N}:H \leftarrow B, fact_{M}: A \leftarrow \}$ with $A \in KB_{k}$, $H \leftarrow B \in P$,
$N=0,..,n$ and $M=0,..m$, which is removed from $KB_{k}$, leading to the reduced program $KB_{k+1} = \{KB_{k} \setminus U^{neg}_{oid}\}$.}
\end{scriptsize}

Applying arbitrary sequences of positive and negative updates leads to a sequence of KB states $KB_{0},..,KB_{k}$ where each state $KB_{i}$ is
defined by either $KB_{i} = KB_{i-1} \cup U^{pos}_{oid_{i}}$ or $KB_{i} = KB_{i-1} \setminus U^{neg}_{oid_{i}}$.

Transactional updates in our rule language are inspired by the serial Horn version of transaction logics (TR) \cite{Bonner1995}.

\begin{scriptsize}
\textbf{Definition}: \textit{(\textbf{Transactional Update}) A transactional update is an update, possibly consisting of several atomic updates, which
must be executed completely or not at all. In case a transactional update fails, i.e., it is only partially executed or violates integrity wrt
to integrity constraints, it will be rolled back otherwise it will be committed. Formally, a transactional
update is defined as follows:\\
\\
$U^{trans}_{oid}:=U^{pos/neg}_{oid_{1}},..,U^{pos/neg}_{oid_{n}} \& \overline{IC}$\\
\\
, where $\overline{IC}=\{IC_{1},..,IC_{m}\}$ is a possibly empty set of integrity constraints which must hold after the update has been
executed. In case an integrity constraint is violated the update is rolled back.}
\end{scriptsize}

A roll back to the previous state before the update means to apply the complement update function on the added or removed modules:

\begin{scriptsize}
\textbf{Definition}:\textit{(\textbf{Rollback of Update}) A transactional update is rolled back by inverting the update
primitive:\\
$KB_{i} = KB_{i+1} \setminus U^{trans}_{oid}$ iff $KB_{i+1} = KB_{i} \cup U^{trans}_{oid}$\\
$KB_{i} = KB_{i+1} \cup U^{trans}_{oid}$ iff exists $KB_{i+1} = KB_{i} \setminus U^{trans}_{oid}$\\}
\end{scriptsize}

Note, that due to the module concept in our language only the transition sequence consisting of the update state oids and the update primitive
needs to be remembered to rollback a sequence of transactional updates.

The declarative semantics of transactional updates is built on the concept of sequences of state transitions $\langle KB_{1},...,KB_{k}\rangle$.
The truths of update are defined on transition paths, i.e. the answer to a query is not determined by the current knowledge base alone, but
depends on the entire transition paths. That is, only paths which return a non-empty answer respectively succeed in case of propositional
queries are considered and their executed transactional updates are committed, whereas paths with empty answer sets are backtracked and the
processed transactional updates within such paths are rolled back to the state of the last backtracking point. A query fails if the set of
answers is empty for every possible transition path. The logical account of transactional execution (derivation with transactional updates) is
given by the concept of executional entailment adapted from TR.

\begin{scriptsize}
\textbf{Definition}: \textit{(\textbf{Executional Entailment}) Let $KB_{0}$ be an initial KB state and $Q$ be a query which might contain free
variables $X_{1},..,X_{n}$ then $KB_{0}, \pi \models Q$, i.e., $Q$ is true in $KB_{0}$, iff there exists a path $\pi= \langle KB_{0}, .. ,
KB_{k} \rangle$ which returns a non-empty answer for all variables in $Q$. $Q$ fails if it returns an empty answer set for every possible
execution paths $\pi_{i}$.}
\end{scriptsize}

That is, a query involves a mapping from sequences of KB states to sets of tuples of ground terms in each state. In case of free queries with
variables several transition paths are considered to produce answers for the query $Q$, then the final state $\overline{KB}$ which becomes the
new knowledge base (state) is the union of all final states of valid transition paths $\pi_{i}$ which entail $Q$. Queries which do not involve
any updates, i.e., which do not consider serial update rules but only LP rules with standard literals without any side effects on the KB, have
an transition path with length $k=0$. In this case, a goal $Q$ is entailed if $KB, \pi \models Q$ and $\pi=\{KB\}$, i.e., there is no state
transition and accordingly executional entailment reduces to standard LP entailment $KB \models Q$.

\section{Integration of Reaction Rules into RuleML}

In this section we propose a reaction rule extension, called Reaction RuleML, to RuleML, as a standard for rule interchange and rule
serialization in XML. Reaction RuleML serves as a platform-independent rule interchange format, and is intended to be translated into
platform-specific, executable rule languages such as our homogenous reaction rule language which we described in the previous sections.

The \textbf{Rule Markup Language} (RuleML) \cite{Boley2006} is a modular, interchangeable rule specification standard to express both forward
(bottom-up) and backward (top-down) rules for deduction, reaction, rewriting, and further inferential-transformational tasks. It is defined by
the Rule Markup Initiative \cite{Boley2000}, an open network of individuals and groups from both industry and academia that was formed to
develop a canonical Web language for rule serialization using XML and for transformation from and to other rule standards/systems. The language family of
RuleML covers the entire rule spectrum, from derivation rules to reaction rules including rule-based event processing and messaging (Reaction
RuleML \cite{Paschke2006c}), as well as verification and transformation rules. In the following, we will briefly summarize the key components of
the Derivation RuleML language (Horn logic layer), and then introduce the Reaction RuleML sublanguage \cite{Paschke2006c} which extends RuleML with
additional language constructs for representing reaction rules and complex event / action messages. The building blocks of Derivation RuleML
are: \cite{Boley2006}

\begin{scriptsize}
\begin{itemize}
    \item Predicates are n-ary relations introduced via an $<Atom>$ element in RuleML. The main terms within an atom are variables $<Var>$ to be instantiated by values when the rules are applied, individual constants  $<Ind>$, data values $<Data>$, and complex expressions $<Expr>$.
    \item Derivation Rules are defined by an $<Implies>$ element which consists of a body part ($<body>$) with one or more atomic conditions connected via $<And>$ or $<Or>$, possibly negated by $<Neg>$ (for classical negation) or $<Naf>$ (for negation as failure), and of a conclusion part ($<head>$) that is implied by the body, where rule application can be in a forward or backward manner.
    \item Facts are stated as atoms deemed to be true: $<Atom>$
    \item Queries $<Query>$ can be proven backward as top-down goals or forward via bottom-up processing, where several goals may be connected within a query, possibly negated.
\end{itemize}
\end{scriptsize}

Besides facts, derivation rules, and queries, RuleML defines further rule types such as integrity constraints and transformation rules
\cite{Boley2006}.

\textbf{Reaction RuleML} \cite{Paschke2006c} is a general, practical, compact and user-friendly XML-serialized sublanguage of RuleML for the
family of reaction rules. It incorporates various kinds of production, action, reaction, and KR temporal/event/action logic rules as well as
(complex) event/action messages into the native RuleML syntax using a system of step-wise extensions. The building blocks of Reaction RuleML
(v0.2) are: \cite{Paschke2006c}

\begin{scriptsize}
\begin{itemize}
    \item One general (reaction) rule form ($<Rule>$) that can be specialized by the selection of the constituent subparts to, e.g. Condition-Action rules (production
    rules), Event-Action rules (trigger rules), Event-Condition-Action rules ...
    \item Three execution styles defined by the attribute $@execution$; default value is "reasoning".
    \begin{itemize}
        \item \textit{active}: 'actively' polls/detects occurred events in global ECA style, e.g. by a ping on a service/system or a query on an internal or external event database
        \item \textit{messaging}: waits for incoming complex event message (inbound) and sends messages (outbound) as actions
        \item \textit{reasoning}: Logical reasoning as e.g., in logic programming (derivation rules) and KR formalisms such as event/\\ action/transition logics (as e.g. in Event Calculus, Situation Calculus, TAL formalizations)
    \end{itemize}
    \item "weak" and "strong" evaluation/execution semantics ("@eval") interpretation which are used to manage the "justification lifecycle" of local inner reaction rules in the derivation/execution process of the outer rules
    \item optional object identifier (\textit{oid})
    \item an optional metadata label (\textit{label}) with e.g. metadata authoring information such as rule name, author, creation date, source ...
    \item an optional set of qualifications (\textit{qualification}) such as a validity value, fuzzy value, a defeasible priority value...
    \item Messages $<Message>$ define inbound or outbound event message
\end{itemize}
\end{scriptsize}

A reaction rule might apply globally as e.g. global ECA rules or locally nested within other reaction or derivation rules as e.g. in the case of
serial messaging reaction rules. The most general syntax of a (reaction) rules is:

\begin{scriptsize}
\begin{verbatim}
<Rule execution="active" eval="strong">
   <oid> <!-- object identifier --> </oid>
   <label> <!-- metadata  --> </label>
   <qualification> <!-- qualifications --> </qualification>
   <on>  <!-- event -->  </on>
   <if>  <!-- condition --> </if>
   <then> <!-- conclusion --> </then>
   <do>  <!--  action -->  </do>
   <after> <!-- postcondition --> </after>
   <else> <!-- else conclusion --> </else>
   <elseDo> <!-- else/alternative action --> </elseDo>
   <elseAfter> <!-- else postcondition --> </elseAfter>
</Rule>
\end{verbatim}
\end{scriptsize}

According to the selected and omitted rule parts a rule specializes, e.g. to a derivation rule (if-then or if-then-else; reasoning style), a
trigger rule (on-do), a production rule (if-do), an ECA rule (on-if-do) and special cases such as ECAP rule (on-if-do-after) or mixed rule types
such as derivation rule with alternative action (if-then-elseDo), e.g. to trigger an update action (add/remove from KB) or send an event message
(e.g. to a log system) in case a query on the if-then derivation rule fails.

Complex event algebra constructs such as $<Sequence>$, $<Disjunction>$,$<Conjunction>$, $<Xor>$, $<Concurrent>$, $<Not>$, $<Any>$,
$<Aperiodic>$, $<Periodic>$ and complex action algebra constructs such as $<Succession>$, $<Choice>$, $<Flow>$, $<Loop>$ can be used in the
event respective action part to define complex event types and complex actions.

The metadata label can be optionally used to add additional metadata using Semantic Web / metadata vocabularies such as Dublin Core. The rule
quantifications such as priorities or validity values form the basis for expressive reasoning formalisms such as defeasible logic, argumentation
semantics, scoped reasoning or fuzzy reasoning (see e.g. \cite{Paschke2007}).\\
In messaging style inbound and outbound messages $<Message>$ are used to interchange events (e.g. queries and answers) and rules:

\begin{scriptsize}
\begin{verbatim}
<Message mode="outbound" directive="ACL:inform">
  <oid> <!-- conversation ID--> </oid>
  <protocol> <!-- transport protocol --> </protocol>
  <sender> <!-- sender agent/service --> </sender>
  <content> <!-- message payload -->  </content>
</Message>
\end{verbatim}
\end{scriptsize}

\begin{scriptsize}
\begin{itemize}
    \item $@mode= inbound|outbound$ -- attribute defining the type of a message
    \item $@directive$ -- attribute defining the pragmatic context of the message, e.g. a FIPA ACL performative
    \item $<oid>$ -- the conversation id used to distinguish multiple conversations and conversation states
    \item $<protocol>$ -- a transport protocol such as HTTP, JMS, SOAP, Jade, Enterprise Service Bus (ESB) ...
    \item $<sender>$$<receiver>$ -- the sender/receiver agent/service of the message
    \item $<content>$ -- message payload transporting a RuleML / Reaction RuleML query, answer or rule base
\end{itemize}
\end{scriptsize}

The directive attribute corresponds to the pragmatic instruction, i.e. the pragmatic characterization of the message context. External
vocabularies defining pragmatic performatives might be used by pointing to their conceptual descriptions. A standard nomenclature of pragmatic
performatives is defined by the Knowledge Query Manipulation Language (KQML) and the FIPA Agent Communication Language (ACL) which defines
several speech act theory-based communicative acts. Other vocabularies such as the normative concepts of Standard Deontic Logic, e.g., to define
action obligations or permissions and prohibitions might be used as well.

The conversation identifier is used to distinguish multiple conversations and conversation states. This allows to associate messages as
follow-up to previously existing conversations, e.g. to implement complex coordination and negotiation protocols. For an overview and
description of several negotiation and coordination protocols see \cite{Paschke2006g}.

The protocol might define lower-level ad-hoc or ESB transport protocols such as HTTP, JMS, SOAP or e.g. agent-oriented communication protocols
such as Jade.

The content of a message might be a query or answer or a larger rule or fact base serialized in RuleML / Reaction RuleML.

RuleMLs' webized typing approach enables the integration of rich external syntactic and semantic domain vocabularies into the compact RuleML
language. For instance, the following example shows a Reaction RuleML message with a IBM Common Base Event (CBE) formalization as payload.

\textbf{Example}
\begin{scriptsize}
\begin{verbatim}
<Message mode="outbound" directive="ACL:inform">
  <oid><Ind>conversation123</Ind></oid>
  <protocol><Ind>esb</Ind></protocol>
  <sender><Ind>server32@lapbichler32</Ind></sender>
  <receiver><Ind>client11@lapbichler32</Ind></receiver>
  <content>
     <And>
        <Atom>
            <oid>
             <Ind type="cbe:CommonBaseEvent">i000000</Ind>
            </oid>
            <Rel use="value" uri="cbe:creationTime"/>
            <Ind type="owlTime:Year">2007</Ind>
            <Ind type="owlTime:Mont">6</Ind>
                    ...
        </Atom>
         ...
        <Atom>
            <oid>
             <Ind type="cbe:CommonBaseEvent">i000000</Ind>
            </oid>
            <Rel use="value" uri="cbe:msg"/>
            <Ind type="xsd:String">Hello World</Ind>
        </Atom>
     </And>
   </content>
</Message>
\end{verbatim}
\end{scriptsize}

The \textbf{RuleML Interface Description Language} (RuleML IDL), as a sublanguage of Reaction RuleML, adopts ideas of interface definition
languages such as CORBA's IDL or Web Services' WSDL. It describes the signature of public rule functions together with their mode and type
declarations, e.g. the public rule heads which can be queried in backward-reasoning style or monitored and matched with patterns in forward
reasoning style.

\emph{Modes} are instantiation patterns of predicates described by mode declarations, i.e. declarations of the intended input-output
constellations of the predicate terms with the following semantics:

\begin{itemize}
    \item "$+$" The term is intended to be input
    \item "$-$" The term is intended to be output
    \item "$?$" The term is arbitrary (input or output)
\end{itemize}

We define modes with an optional attribute $@mode$ which is added to terms in addition to the $@type$ attribute. By default the mode is
arbitrary $"?"$. An interface description is represented by the function $interface(QueryLiteral, Description)$, where\\
$QueryLiteral$ is the signature of the publicly accessible literal. Thus, the interface definition for the function $add(Result,Arg1,$\\
$Arg2)$ with the modes $add(-,+,+)$ is as follows:

\textbf{Example}
\begin{scriptsize}
\begin{verbatim}
<Atom>
    <Rel>interface</Rel>
    <Expr>
        <Fun>add</Fun>
        <Var type="java:java.lang.Integer" mode="-">Result</Var>
        <Var type="java:java.lang.Integer" mode="+">Arg1</Var>
        <Var type="java:java.lang.Integer" mode="+">Arg2</Var>
    <Expr>
    <Ind>Definition of the add function which takes two Java
        integer values as input and returns the Integer result
        value</Ind>
</Atom>
\end{verbatim}
\end{scriptsize}

\section{Rule-based CEP Middleware}
In this section we will implement the main components of the rule-based middleware which makes use of the homogenous reaction rule language. As
in a model driven architecture (MDA) the middleware distinguishes:

\begin{scriptsize}
\begin{itemize}
    \item a platform specific model (PSM) which encodes the rule statements in the language of a specific execution environment
    \item a platform independent model (PIM) which represents the rules in a common (standardized) interchange format (e.g. a markup language)
    \item a computational independent model (CIM) with rules in a natural or visual language.
\end{itemize}
\end{scriptsize}

We focus on the technical aspects of the middleware and on the machine-to-machine communication between automated rule inference services
deployed on the middleware. The homogenous reaction rule language is used on the PSM layer and RuleML/Reaction RuleML on the PIM layer. For an
visual RuleML/Reaction RuleML editor on the CIM layer see \cite{Paschke2007}. Figure \ref{Middleware} illustrates the architecture of the
middleware.

\begin{figure}
\centering
\includegraphics[width=8cm]{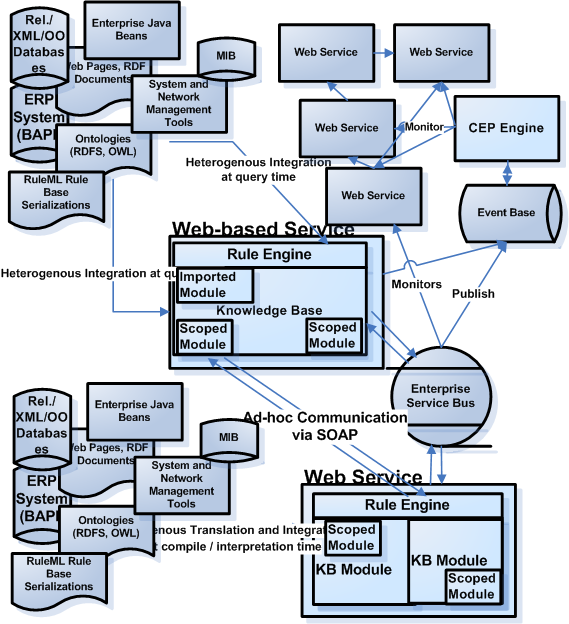}
\caption{Rule-based CEP Middleware} \label{Middleware}
\end{figure}

Several rule engines (e.g. Prova \cite{Kozlenkov2006} or other rule execution environments) might be deployed as distributed web-based services.
Each service dynamically imports or pre-compiles and loads the distributed rule bases (RuleML scripts published on the web or stored locally)
which implement the decision and reaction logic. External data from data sources such as Web resources or relational databases and external
enterprise application tools, web services and object representations can be directly integrated during runtime or by translation during compile
time by the expressive homogenous and heterogenous integration interfaces of the rule engine or via publish-subscribe and monitoring processes
using the ESB and event messages. The ESB is used as object broker for the web-based inference services and as asynchronous messaging middleware
between the services. Different transport protocols such as JMS, HTTP or SOAP (or Rest) can be used to transport rule sets, queries and answers
as payload of event messages between the internal inference services deployed on the ESB endpoints or external systems and applications.
RuleML/Reaction RuleML is used as common rule interchange format between the services.

\subsection{Enterprise Service Bus as Communication Middleware}

Our middleware is based on the Mule open-source ESB \cite{Mule2006} in order to handle message-based interactions between the deployed
rule-based services and with other external applications, tools and services using disparate complex event processing (CEP) technologies,
transports and protocols. The ESB allows to deploy the rule engines  as highly distributable rule inference services installed as Web-based
endpoints on the Mule object broker and supports the Reaction RuleML based communication between them. That is, the ESB provides a highly
scalable and flexible application messaging framework to communicate synchronously but also asynchronously with external and internal services.

Mule is a messaging platform based on ideas from ESB architectures, but goes beyond the typical definition of an ESB as a transit system for
carrying data between applications by providing a distributable object broker to manage all sorts of service components. A transport provider
enables Mule components to send and receive information over a particular protocol, repository messaging or other technology. Mule supports a
great variety of transport protocols such as JMS, HTTP, SOAP, TCP. Autonomous components such as Java Beans or components from other frameworks
are managed within the object broker and configured to exchange inbound and outbound event messages through registered routers to the
components' endpoint addresses. The three processing modes of Mule are \cite{Mule2006}:

\begin{scriptsize}
\begin{itemize}
    \item Asynchronously: many events can be processed by the same component at a time in various threads. When the Mule server is running
    asynchronously instances of a component run in various threads all accepting incoming events, though the event will only be processed by one
    instance of the component.
    \item Synchronously: when a Component receives an event in this mode the whole request is executed in a single
    thread
    \item Request-Response: this allows for a Component to make a specific request for an event and wait for a specified time to get a
    response back
\end{itemize}
\end{scriptsize}

The object broker follows the Staged Event Driven Architecture (SEDA) pattern \cite{Welsh2001}. The basic approach of SEDA is to decomposes a
complex, event-driven application into a set of stages connected by queues. This design decouples event and thread scheduling from application
logic and avoids the high overhead associated with thread-based concurrency models. That is, SEDA supports massive concurrency demands on
Web-based services and provides a highly scalable approach for asynchronous communication. Figure \ref{MuleProva} shows a simplified breakdown
of the Mule integration.

\begin{figure}
    \begin{center}
        \includegraphics[width=6cm]{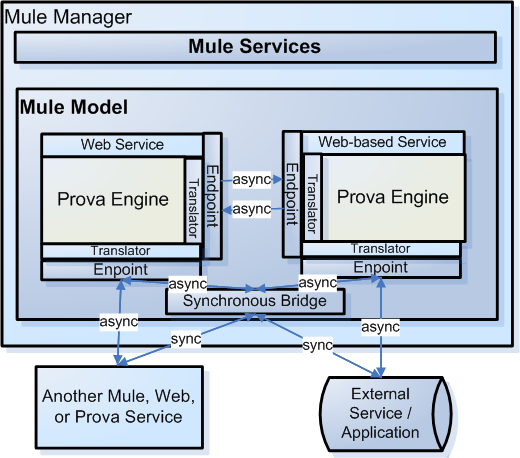}
    \end{center}
    \caption{Distributed Inference Services on an ESB}\label{MuleProva}
\end{figure}

Several inference services which at their core run a rule engine are installed as Mule components which listen at configured endpoints, e.g.,
JMS message endpoints, HTTP ports, SOAP server/client addresses or JDBC database interfaces. Reaction RuleML is used as a common platform
independent rule interchange format between the agents (and possible other rule execution / inference services). Translator services are used to
translate inbound and outbound messages from PIM Reaction RuleML into the PSM rule engines execution syntaxes and vice versa. XSLTs and ANTLR
based translator services are provided as Web forms, HTTP services and SOAP Web services on the Reaction RuleML Web page \cite{Paschke2006c}.

The large variety of transport protocols provided by Mule can be used to transport the messages to the registered endpoints or external
application objects. Usually JMS is used for the internal communication between distributed agent instances, while HTTP and SOAP is used to
access external Web services. The usual processing style is asynchronous using SEDA event queues. However, sometimes synchronous communication
is needed. For instance, to handle communication with external synchronous HTTP clients such as Web browsers where requests, e.g. by a Web from,
are send through a synchronous channel. In this case the implemented synchronous bridge component dispatches the requests into the asynchronous
messaging framework and collects all answers from the internal service nodes, while keeping the synchronous channel with the external service
open. After all asynchronous answers have been collected they are send back to the still connected external service.

\subsection{Platform-dependent Rule Engines as Execution Environments}

Each web-based inference service might run one or more arbitrary rule engines to execute the interchanged rules and events and derive answers on
requests. In this subsection we will introduce Prova \cite{Kozlenkov2006}, a highly expressive open-source Semantic Web rule engine which
supports complex reaction rule-based workflows, decision logic and dynamic access to external data sources and Java objects.

Prova follows the spirit and design of the recent W3C Semantic Web initiative and combines declarative rules, ontologies and inference with
dynamic object-oriented Java API calls and access to external data sources such as relational databases or enterprise applications and IT
services. One of the key advantages of Prova is its elegant separation of logic, data access, and computation and its tight integration of Java
and Semantic Web technologies. It includes numerous expressive features and logic formalisms such as:

\begin{scriptsize}
\begin{itemize}
    \item Easy to use and learn ISO Prolog related scripting syntax
    \item Well-founded Semantics for Extended Logic Programs with defeasible conflict resolution and linear goal memoization
    \item Order-sorted polymorphic type systems compatible with Java and Semantic Web ontology languages RDF/RDFS and OWL
    \item Seamless integration of dynamic Java API invocations
    \item External data access by e.g., SQL, XQuery, RDF triple queries, SPARQL
    \item Meta-data annotated modular rule sets with expressive transactional updates, Web imports, constructive views and scoped reasoning for
    distributed rule bases in open environment such as the Web
    \item Verification, Validation and Integrity tests by integrity constraints and test cases
    \item Messaging reaction rules for workflow like communication patterns based on the Prova Agent Architecture
    \item Global reaction rules based on the ECA approach
    \item Rich libraries and built-ins for e.g. math, date, time, string, interval, list functions
\end{itemize}
\end{scriptsize}

For a detailed description of the syntax, semantics and implementation see \cite{Paschke2007,Kozlenkov2006}.

\section{Conclusion}
Flexibility in dynamically composing new business processes and integrating heterogenous information systems (HIS), enabling ad-hoc cooperations,
is one of the main aims of the recent service oriented computing (SOC) paradigm. The vision is to build large-scale service supply chains
(a.k.a. business services networks) which enable enterprises to define and execute Web Services based transactions and business processes across
multiple business entities and domain boundaries using standardized (Web) protocols.

In this paper we propose a logic-based homogenous reaction rule language and a rule and event-based middleware which combine technologies from
declarative rule-based programming with enterprise application technologies for CEP and SOC. The rule-based approach
follows the separation-of-concerns principle and addresses imperatives businesses face today: Deploy the business logic in a
service-oriented declarative way, effectively detect, communicate and adequately react to occurred complex events in the business and service
management processes, change the decision and behavioral business logic which underpins their applications and service offerings in order to
adapt to a flexible business environment (e.g. on-demand/utility computing), and to overcome the restrictive nature of slow IT change cycles.
Our rule-based approach has the potential to profoundly change the way IT services are used and collaborate in business processes. Akin to
multi-agent systems (MAS), the rule-based logic layer allows for intelligent semi-autonomous decisions and reactions which are necessary, e.g.,
for IT service management (ITSM) processes such as service level management (SLM), change management, availability management and business
process management (BPM) such as service execution in workflow-like business processes and business activity monitoring (BAM). We have
successfully demonstrated the usability and adequacy of our rule-based approach in various domains of research and industry use cases such as
for ITSM and BAM (see RBSLA project \cite{Paschke2006}), Semantic Search (see GoPubMed project \cite{Kozlenkov2004}), and Semantic Web-based
virtual organizations and collaboration (see Rule Responder project \cite{Paschke2007a}).

We argue that our rule-based design artifact (service-oriented CEP middleware + declarative rule language and rule engine) is a more
expressive and adequate tool to describe complex event processing logic and business service executions than, e.g., the too much focused event
correlation engines or the very restricted "if-activity" rules in BPEL (BPEL 2.0) which are only based on simple material implications and
not on the precise and expressive semantics of logic programming and non-monotonic reasoning formalisms. The clear logical semantics of our
approach ensures traceability, verifiability and reliability of derived results and reactions. Ultimately, our rule-based design artifact might
help to put the vision of highly flexible and adaptive business services networks with support for CEP, ITSM, BAM and BPM into large-scale practice.

\bibliographystyle{abbrv}
\bibliography{sigproc}  
\end{small}

\end{document}